\documentclass[letterpaper, 10 pt, conference]{ieeeconf}  
\usepackage[utf8]{inputenc}

\IEEEoverridecommandlockouts                              

\usepackage{graphics} 
\usepackage{epsfig} 
\usepackage{mathptmx} 
\usepackage{times} 
\usepackage{amsmath} 
\usepackage{amssymb}  
\usepackage{needspace}

\usepackage{xcolor}
\usepackage{hyperref}
\definecolor{blue}{RGB}{0, 128, 128}
\hypersetup{
  colorlinks=true,     
  linkcolor=blue,      
  urlcolor=blue,       
  filecolor=blue       
}

\usepackage{soul}
\usepackage{cite}
\usepackage{booktabs}
\usepackage[font=small,labelfont=bf,labelsep=period]{caption}
\usepackage{multirow}
\usepackage{float}

\title{\LARGE \bf
IDfRA: Self-Verification for Iterative Design in Robotic Assembly}

\author{Nishka Khendry$^{1}$ and Christos Margadji$^{1}$ and Sebastian W. Pattinson$^{1}$\\[0.5em]
\textit{ Accepted at the 2026 IEEE International Conference on Robotics and Automation (ICRA)}
\thanks{© 2026 IEEE.  Personal use of this material is permitted.  Permission from IEEE must be obtained for all other uses, in any current or future media, including reprinting/republishing this material for advertising or promotional purposes, creating new collective works, for resale or redistribution to servers or lists, or reuse of any copyrighted component of this work in other works. \textit{DOI: 10.1109/ICRA57385.2026.11697014}}
\thanks{Codebase: {\href{https://github.com/cam-cambridge/open-ended-assembly}{https://github.com/cam-cambridge/open-ended-assembly}}}
\thanks{This work was supported by the UKRI Engineering
and Physical Sciences Research Council award EP/N509620/1.}
\thanks{$^{1}$Institute for Manufacturing, Department of Engineering, University of Cambridge
        {\tt\small \{nk680,cm2161,swp29\}@cam.ac.uk}}%
}

\begin{document}

\maketitle

\thispagestyle{empty}
\pagestyle{empty}

\begin{abstract}
Design for Robotic Assembly (DfRA) remains largely dependent on manual planning and heuristic simulation, limiting scalability and robustness in complex industrial settings. Although large language models (LLMs) show promise for semantic reasoning and task planning, most approaches remain tightly coupled to pre-built simulators that assume an accurate world model.
We introduce Iterative Design for Robotic Assembly (IDfRA), a closed-loop framework that combines an LLM for plan generation with a vision–language model (VLM) for execution assessment. Given a target structure and a partial environmental signature, the LLM proposes an assembly plan, the robot executes it once at test time, and the VLM evaluates the resulting state to provide feedback for replanning. Through this iterative planning–execution–verification loop, the system progressively improves semantic fidelity and physical feasibility.
Crucially, IDfRA does not require an accurate a priori world model before deployment. Instead, physical constraints are discovered online through interaction, enabling adaptation to under-specified environments.
Empirical evaluation demonstrates that IDfRA attains 73.3\% top-1 accuracy in semantic recognisability, surpassing the baseline on this metric. Moreover, the resulting assembly plans exhibit robust physical feasibility, achieving an overall 86.9\% construction success rate, with design quality improving across iterations, albeit not always monotonically. Pairwise human evaluation further corroborates the advantages of IDfRA relative to alternative approaches. By integrating self-verification with context-aware adaptation, the framework evidences strong potential for deployment in unstructured manufacturing scenarios.
\end{abstract}


\section{Introduction}
Robotic assembly involves using robots in manufacturing to automatically assemble components into finished products. While traditionally limited to repetitive tasks for efficiency \cite{patent}, manufacturing is shifting toward flexible, adaptive systems that operate in dynamic, unstructured environments \cite{rob_assemb_review, sustainable}. This transition has driven interest in Design for Robotic Assembly (DfRA), which involves concurrent design of both products and the robotic systems that assemble them \cite{2005book}. However, conventional DfRA remains largely manual and template-based \cite{trad_dfra1, trad_dfra2, trad_dfra3}, constrained by vast design spaces and reliance on experience-based heuristics -- often resulting in suboptimal outcomes \cite{manual_dfra_bad}.

\begin{figure} 
    \centering
    \includegraphics[width=\linewidth]{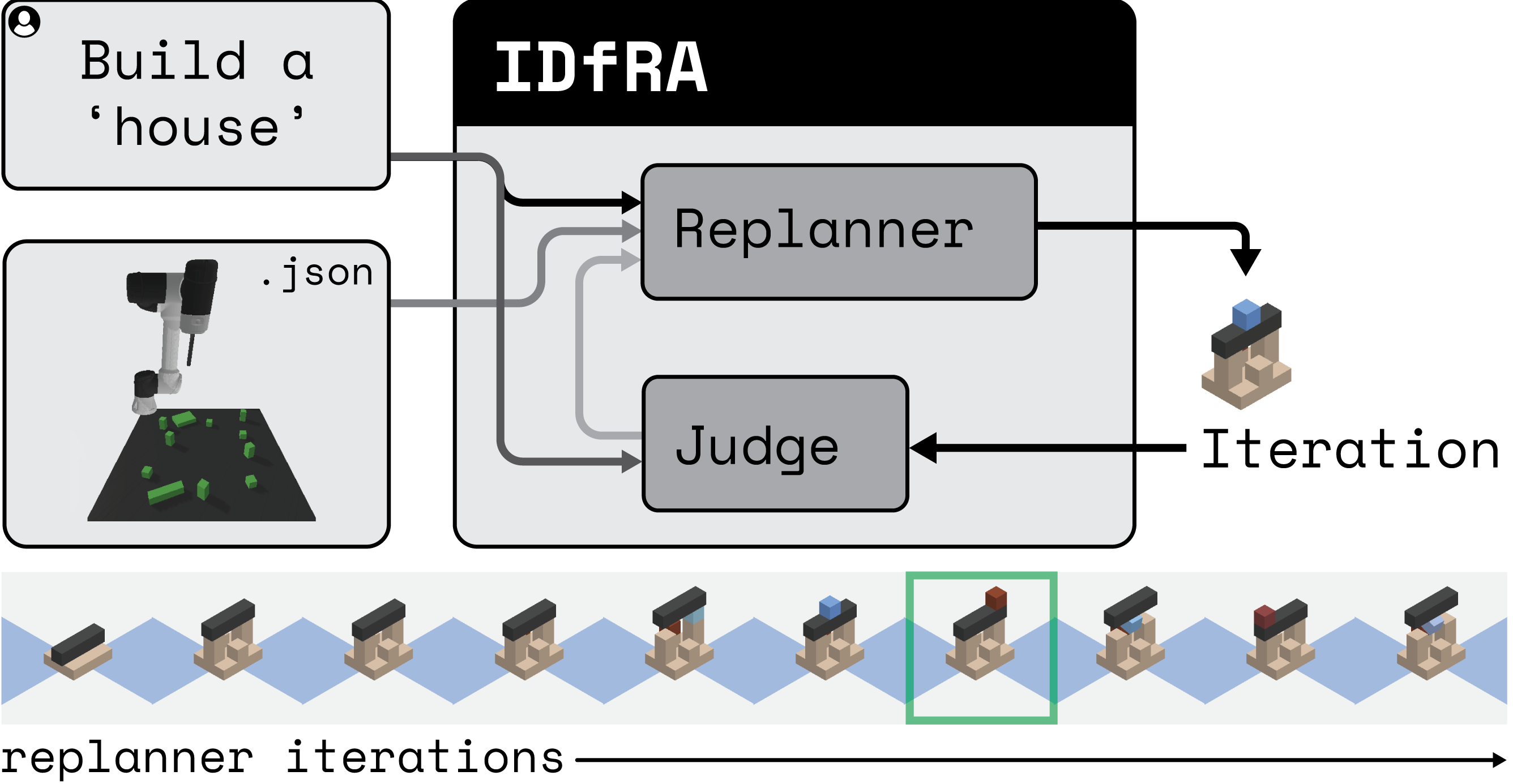}
    \caption{\textbf{The overall IDfRA framework.} \textbf{Top}: Given a target assembly (e.g., ``house") and a signature of the environment in .json format, IDfRA iteratively generates a detailed assembly plan specifying each available block's position and pose. \textbf{Bottom}: Renders of the 10  iterations of the ``house" assembly.}
    \label{fig:intro_diag}

    \vspace{4mm}
    \includegraphics[width=\linewidth]{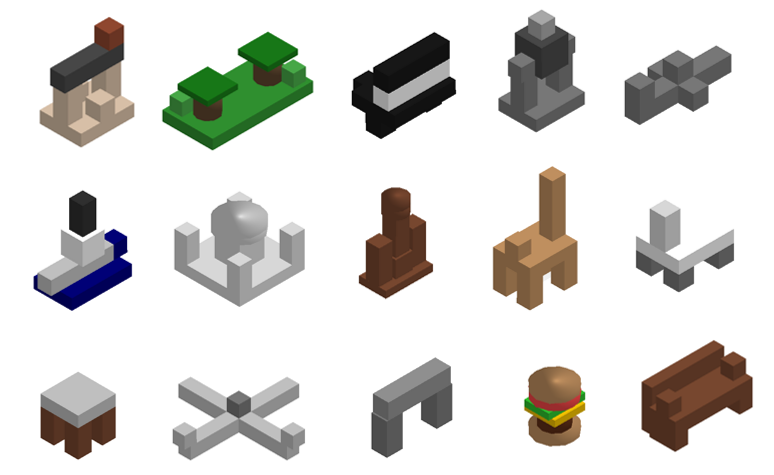}
    \caption{\textbf{15 representative assemblies} generated by IDfRA. Top left to bottom right: house, jungle, piano, robot, shark, ship, Taj Mahal, Eiffel Tower, giraffe, sheep, dining table set, ceiling fan, bridge, burger, couch.}
\end{figure}

To automate DfRA there has been focus on Assembly Sequence Planning (ASP), which optimises the order of pre-specified assembly steps. 
Notable examples include a transformer-based model that derives assembly sequences from a target blueprint \cite{past}, and a constraint-based tree search to generate feasible sequences automatically \cite{asap}. However, such methods address only a subset of DfRA and lack adaptability to dynamic contexts. 

Recently, large language and by extension vision-language models (LLMs and VLMs) have shown strong capabilities in semantic reasoning and high-level task planning \cite{gpt1, gpt2, saycan, tp1, tp2, tp3}, including promising results in diverse manufacturing simulations \cite{embod_manuf}. Leveraging this, \cite{bloxnet} proposed the first VLM-driven DfRA system. However, their method selects one from multiple candidate designs generated in parallel, which precludes dynamic, incremental enhancement of design features or learning from past mistakes.

As manufacturing shifts toward flexibility, there is a need for DfRA systems that iteratively improve through feedback. Such adaptability enables learning from previous mistakes and context-aware design refinement, which are crucial in variable industrial environments \cite{c1, MARGADJI2024104271}. To address this, we introduce IDfRA, a framework for Iterative DfRA, which integrates self-verification and re-planning to enable such adaptability; this creates scope to optimally accommodate changing assembly contexts, such as variations in available components, site conditions, or manufacturing tolerances.

IDfRA employs self-verification, a robotic system’s ability to assess its own actions and outcomes, as its feedback mechanism. Self-assessment is vital in uncertain or safety-critical environments \cite{sv_needed1}, and has been shown to improve performance by enabling iterative refinement of task plans in both robotics \cite{phoenix, copal, monologue} and other domains \cite{voyager}. IDfRA  addresses three core needs in modern DfRA: \textit{adaptability to dynamic contexts} via iterative refinement, \textit{safety} through embedded self-assessment, and \textit{efficiency} via end-to-end automation. The contributions of this work are:
\begin{itemize}
    \item IDfRA, which to our knowledge, is the first fully automated framework combining VLM-based self-verification with LLM-based re-planning for iterative assembly design improvement.
    \item  A systematic evaluation showing that IDfRA generates semantically meaningful, physically feasible designs that evolve through feedback, providing insights into foundation model reasoning within DfRA. 
\end{itemize}

\begin{figure}[t] 
    \centering
    
    \label{fig:all_15}
\end{figure}

\section{Related work}
\subsection{Robotic Assembly}\label{ra_rel_work}
Robotic systems are now widely integrated into industrial manufacturing \cite{ifr_report}, particularly in automotive and aerospace, where they have significantly enhanced production efficiency \cite{rob_assemb_review}. Consequently, Design for Robotic Assembly (DfRA) has emerged as a foundational concept. 

Early methods for Design for Assembly (later extended to robotics) optimised assembly processes to minimise cost \cite{dfa_ori_booth, dfa_ori_hitachi, dfa_ori_lucas} via rule-based and manual strategies. Their limitations in quantifying executability motivated a shift towards physics-based reasoning, which catalysed widespread adoption of Computer-Aided Design (CAD) and Computer-Aided Manufacturing (CAM) software for visualisation and manual evaluation of components and assemblies \cite{cad1, cad2,cad3}.  Contemporary CAD and simulation-based tools, integrated with updated DfRA frameworks derived from the original models \cite{dfa_ori_booth, dfa_ori_hitachi, dfa_ori_lucas} remain the industry standard. Although these approaches advance partial automation of DfRA, they still necessitate close collaboration with engineers throughout the process. Manual DfRA is hindered by a vast solution space, which often leads to suboptimal designs and heavy reliance on designers’ procedural knowledge and intuition \cite{manual_dfra_bad}. Thus, greater automation could significantly improve efficiency and design quality.

Recent shifts in manufacturing towards flexibility and adaptability \cite{flexible_manuf} have amplified interest in automating diverse assembly tasks in unstructured environments \cite{rob_assemb_review, sustainable}. Combined with the limitations of manual methods, this has driven application of ML to robotic assembly, primarily to ASP while DfRA remains largely manual. For example, much ASP research \cite{removability, srsa, asap, mdp_asp, rasp} builds on Assembly-by-Disassembly (AbD), which analyses disassembly steps to inform assembly plans \cite{asp_ori_know}. Meanwhile, LLMs and VLMs have been employed to decode IKEA instruction manuals \cite{manual2skill} and engineering documentation \cite{designQA}, manipulate CAD sketches \cite{cad_llm}, decompose high-level construction tasks \cite{robogpt}, and generate robot code from assembly specifications \cite{code_assem_gpt}. These methods demonstrate the significant potential of large pre-trained models in robotic assembly. However, prior ASP work focuses on sequencing predefined designs rather than generating new, physically valid assemblies from scratch.

While BloxNet \cite{bloxnet} successfully introduced automated DfRA via VLM-based selection, its architecture has key bottlenecks. Its reliance on physics simulation for internal stability correction limits its real-world applicability, as simulation models often struggle with the inaccuracies of physical assembly \cite{sim2real2}. Our framework, conversely, treats simulation only as a final execution placeholder, keeping the reasoning process independent of simulation biases. Furthermore, BloxNet’s minimal verification loop (three iterations) and parallel generation strategy preclude the system from learning from previous mistakes. We propose IDfRA, which utilizes an iterative self-verification loop to enable dynamic, context-aware design refinement. Ultimately, we focus on the core challenge identified by \cite{bloxnet}: the reliable application of LLMs to embodied robotics.

\begin{figure}[t] 
\vspace{2mm}
    \centering
    \includegraphics[width=1\linewidth]{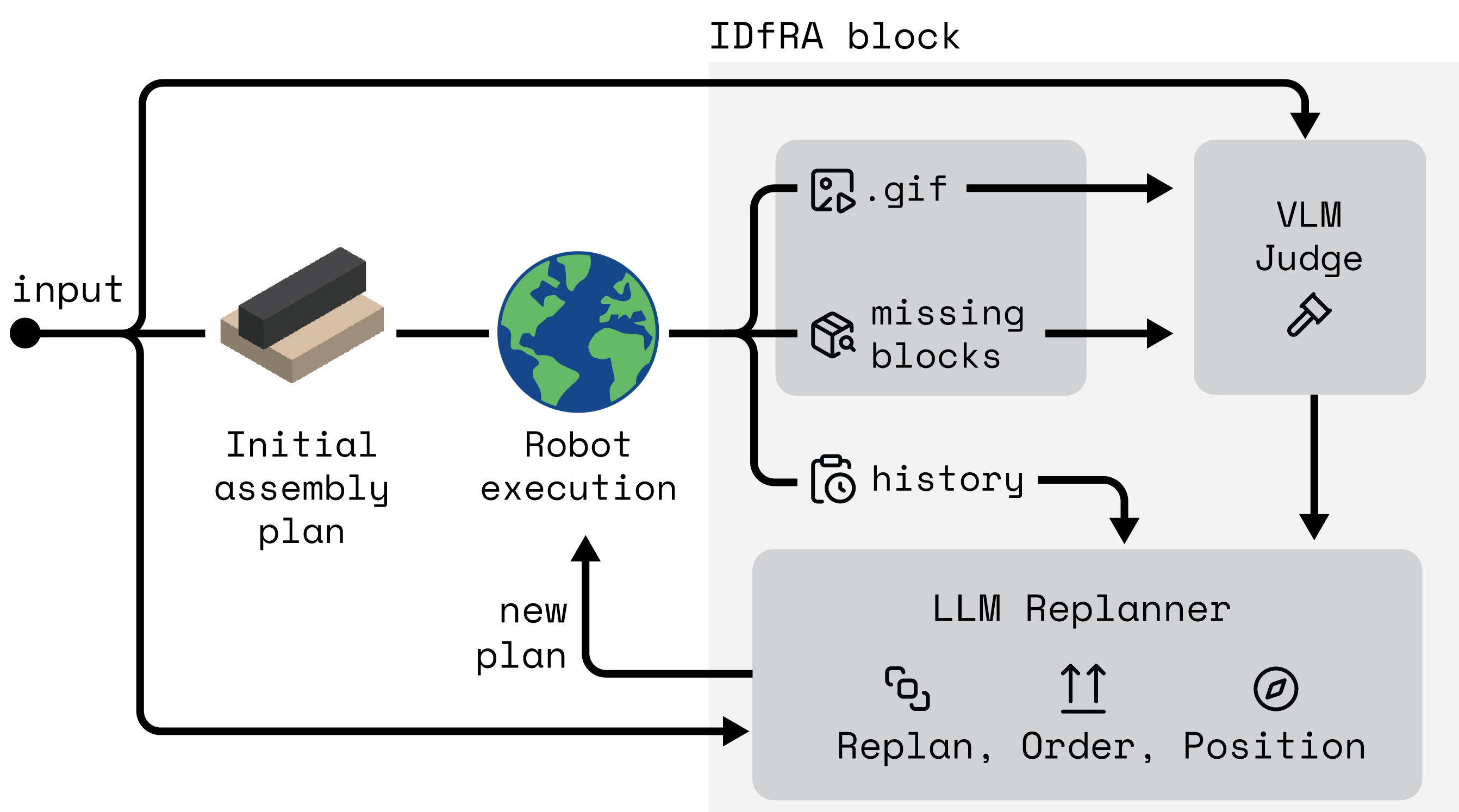}
    \caption{\textbf{IDfRA system architecture.} A GIF of the initial design execution, any missing blocks, and the original inputs are passed to the \textit{VLM Judge}. Based on the \textit{Judge}'s feedback and previous plans, the \textit{Replanner} produces an updated design which is then executed. This process repeats for 10 iterations and a \textit{VLM Selector} chooses the final assembly.}
    \label{fig:sys_arch}
\end{figure}

\subsection{Robotic Planning and Self-Verification with Large Pre-Trained Models}\label{related_lpts}

LLMs offer strong potential for robotic planning due to their broad semantic understanding and creativity \cite{toolsmith}, Yet, they often fail to generate feasible plans in a single pass, and perform far better when combined with external verification and feedback \cite{modulo}. This insight has driven a surge of research into iterative planning pipelines.

Several works employ execution-derived signals, such as simulator outputs or code errors, as sources of feedback \cite{copal, robotgpt_manip, voyager}. In contrast, others use language-based correction or guidance to enhance iterative planning such as via knowledge bases of compliant plans \cite{introspective} and predefined task guidelines \cite{itp}. Furthermore, \cite{monologue} demonstrates the value of combining diverse feedback sources, such as success detection and scene description, and reports significant improvement in task completion across multiple robot environments. Collectively, these studies demonstrate the value of feedback-driven refinement and the promise of LLMs/VLMs for iterative assembly design.

\section{Methods}

\subsection{Task Definition}\label{task_def}
We adopt the Generative DfRA task formulation introduced by BloxNet \cite{bloxnet}. Overall system inputs include a target structure name (e.g., ``house") and a JSON-formatted list of available blocks (including their dimensions, quantities, and shapes: cuboidal or cylindrical).
The goal is to generate a complete assembly plan, specifying for each block its shape, semantic name, colour, 3D centre coordinate (X, Y, Z), and yaw angle. 

\subsection{System Architecture}\label{sys_arch}
While we adopt the same task definition as BloxNet, our approach addresses its key limitations (Section~\ref{ra_rel_work}). Design generation is guided by three objectives: (1) semantic resemblance to the target, (2) physical feasibility for construction by the simulated robot, and (3) iterative enhancement of design features via feedback.
The overall architecture and execution flow of IDfRA is detailed in Fig~\ref{fig:sys_arch}. The process begins with a single candidate design generated by BloxNet, with physics-based stability correction disabled to allow refinement of potentially unstable configurations. Object tracking is handled by PyBullet, while missing blocks are identified by comparing the assembly plan against the available set using Python. Assuming that these functions can be replicated in physical systems via vision-based localisation is reasonable--given recent advancements in generalist agents capable of accurate object tracking and pick-and-place \cite{okrobot, rt-1}.
The IDfRA framework consists of four key components:

\begin{figure*}[t] 
\vspace{2mm}
    \centering
    \includegraphics[width=\textwidth]{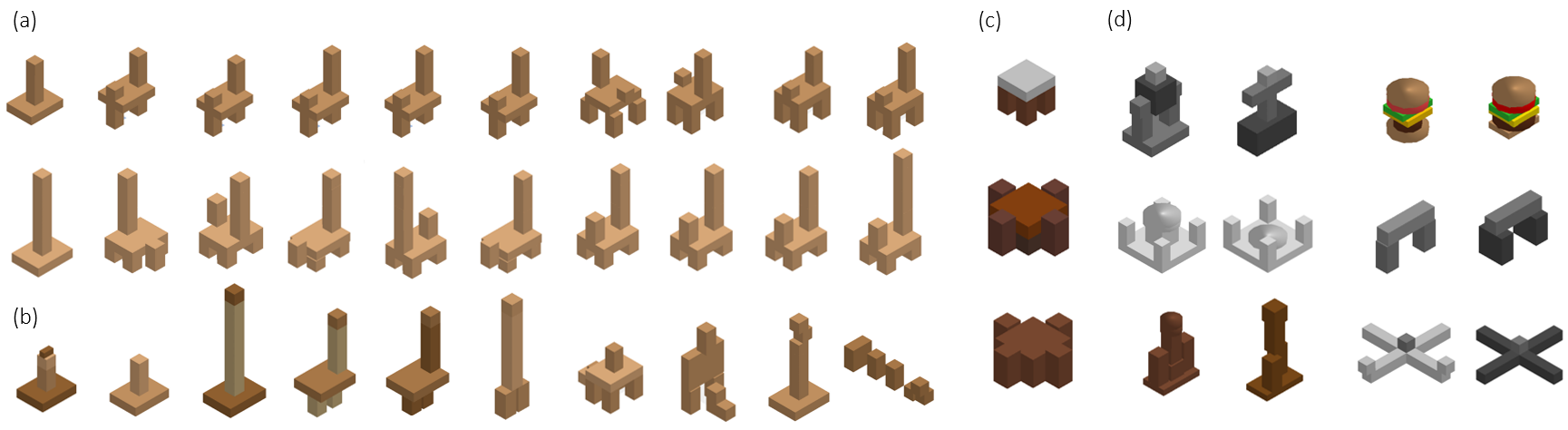}
    \caption{\textbf{IDfRA design generation.} (a) Multiple runs of IDfRA -- each comprising 10 iterations -- produce different design evolutions given the same blocks and input ``giraffe''. (b) 10 candidate designs produced in parallel by BloxNet -- each design is produced independently without influence from the previous one. (c) Designs generated by multiple full runs of IDfRA for ``dining table set". (d) Designs generated by IDfRA (left) and BloxNet (right) for ``robot", ``Taj Mahal", ``Eiffel Tower", ``burger", ``bridge", ``ceiling fan".}
    \label{fig:qual}
\end{figure*}

\subsubsection{LLM-Based Planning Module (\textit{Planner/Replanner})}

The planning module--referred to as the \textit{Planner} in the first iteration and the \textit{Replanner} in subsequent iterations--shares an identical architecture throughout. It consists of three coordinated LLM instances arranged in series to transform high-level intent into an executable assembly plan as in~\cite{bloxnet, copal}. It receives the original task specification and, from the second iteration onward, the latest \textit{Judge} feedback together with a record of previous plans. The output is a JSON assembly plan specifying block dimensions and target block poses (3D centre coordinates, and yaw angles), i.e., where each block should be placed.

First, \textit{Plan} generates a high-level design (or in later iterations--referred to as \textit{Replan}--revises the design) while respecting block availability constraints. It specifies each block’s semantic role, dimensions, colour (for visualisation), and relative placement. Second, \textit{Order} enforces bottom-up feasibility by correcting illogical construction sequences. Finally, \textit{Position} converts the structured design into explicit spatial coordinates (X, Y, Z coordinates and yaw angles). \textit{Replan} simplifies this step by allowing block dimension switching for reorientation, as in~\cite{bloxnet}, enabling \textit{Position} to focus on spatial layout. This stage required careful prompt engineering to minimise overhangs
and enhance structural stability--employing strategies such as removing ambiguity, adding negative examples, and providing detailed instructions that guide positioning based on adjacent blocks. This use of few-shot in-context learning enabled precise placement without relying on a pre-built geometric world model.

\subsubsection{Robot Assembly Execution Module}\label{robot_exec}

The generated plan is executed by a simulated robotic arm in a tabletop environment. Each block is placed via pick-and-place actions with inverse kinematics control. Required reorientations are handled programmatically through 90° roll or pitch adjustments before placement. A suction-based end effector minimises disturbance during stacking.

\subsubsection{VLM-Based Evaluation Module (\textit{Judge})}

The \textit{Judge}, implemented as a single VLM, evaluates each assembly attempt. It receives a GIF of the executed structure together with  structured metadata (block inventory and target specification) describing the generated plan and missing blocks. Using these, the model infers the semantic roles of blocks from visual input and returns structured feedback.

The assessment includes: (1) violations of availability constraints, (2) potential stability concerns, (3) degree of semantic resemblance to the target, and (4) concrete design improvement suggestions. Explicit block availability analysis was included in the prompt to mitigate LLMs' tendency to hallucinate unavailable components – an issue also noted by~\cite{bloxnet}. Feedback prioritises semantic scene understanding, which current VLMs perform more reliably than detailed physical tracking. 

\subsubsection{VLM-Based Selector Module}

After ten refinement iterations, images are rendered for
each assembly by spawning blocks at the generated poses
under gravity and rigid body dynamics. Designs containing missing components are discarded. The remaining candidates undergo pairwise comparison using a VLM, which selects the more structurally stable and semantically faithful design at each stage. This knockout process continues until a single final assembly is chosen.

\subsection{Implementation Details}
The task is executed in a simulated tabletop environment using a Universal Robot UR5e arm and custom suction gripper, both instantiated and controlled using \href{https://pybullet.org/wordpress/}{PyBullet}. Simulation parameters include: uniform object density \(1000~\mathrm{kg/m^3}\), lateral friction coefficient 0.5, spinning friction coefficient 0.2, gravity \(-9.81~\mathrm{m/s^2}\). At each reset, block positions are randomised outside a configurable central region to reserve space for assembly, with minimum spacing to prevent overlap. The simulation acts as an interactive test-time environment: assembly attempts are rendered as GIFs that reflect the realised physical outcome, serving as the core feedback signal for subsequent refinement.

All LLM and VLM instances in this study utilize GPT-4o, accessed via the commercially available API. Consequently, system performance, particularly with respect to outcome assessment, is inherently bounded by the capabilities and limitations of the underlying language model. However, the framework itself is model-agnostic and as foundation models continue to scale and improve, more capable models can be substituted directly without requiring modifications to the overall architecture. JSON is used as the main input at all times, as it is the de facto standard for structured data exchange \cite{gpt_json}. Images and GIFs were encoded based on the provided protocols. Temperatures were tuned to balance creativity and consistency at different stages of the pipeline: 0.4 (\textit{Judge}), 0.5 (\textit{Replan}, \textit{Order}), 0.25 (\textit{Position}), and 0 (\textit{Selector}).

\begin{figure}[t] 
    \centering
    \includegraphics[width=\linewidth]{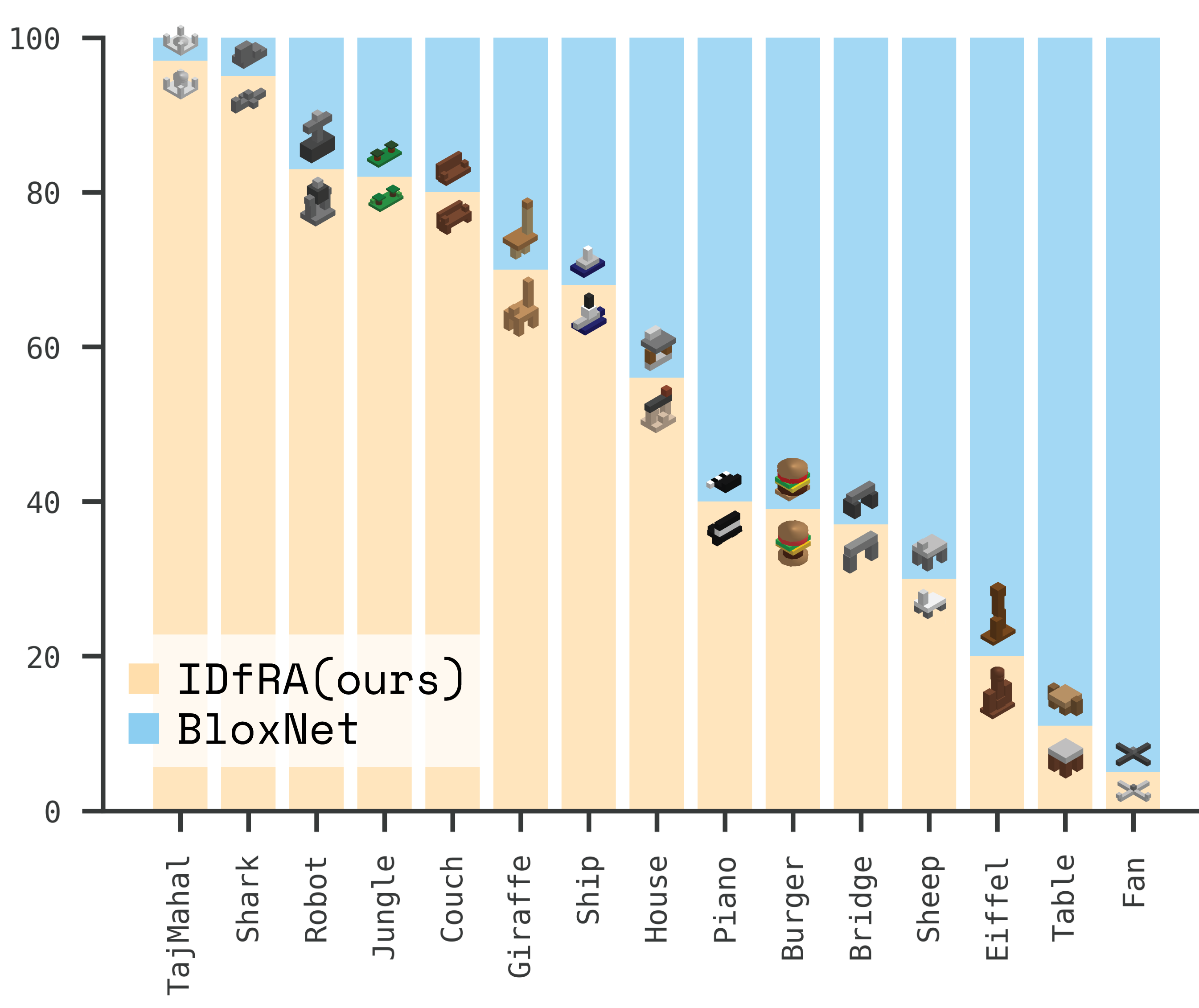}
    \caption{\textbf{Human evaluation of final designs}. Pairwise voting scores from 100 survey responses comparing IDfRA vs BloxNet across 15 assemblies (best viewed zoomed-in.)}
    \label{fig:human_eval}
\end{figure}

\section{Experiments}
We evaluate IDfRA on the three objectives defined in Section~\ref{sys_arch}: semantic recognisability, physical feasibility, and iterative improvement. A representative set of 15 assembly designs was selected to balance structural complexity and tractability: (1) each target structure is sufficiently intricate to support iterative refinement and to incorporate multiple components to achieve semantic resemblance, yet simple enough to be approximated using a limited block set, (2) the structures span varying levels of complexity to illustrate system flexibility.

For each target, a custom block set was prepared to: (1) provide the minimum blocks required for the target structure, (2) include surplus block types to allow creative variation rather than forcing a single design, and (3) remain within the tabletop workspace limits to support physical mappability. All JSON-formatted block sets (including dimensions and quantities) are available in the codebase.

Due to the stochastic nature of autoregressive decoding in VLMs, multiple IDfRA runs produce different design evolutions (Fig.~\ref{fig:qual}a), some outperforming others (Fig.~\ref{fig:qual}c). This variability also highlights IDfRA’s ability to adaptively refine plans based on context and feedback. To ensure fair comparability with baseline BloxNet \cite{bloxnet}, and given the absence of detailed information regarding the exact block sets and number of runs used in its reported results, we ran both methods once using identical inputs (i.e., the target structure name and block set). Each method automatically generated its best designs, and we conducted two quantitative analyses to directly compare their performance. In addition, two analyses were conducted independently on IDfRA's outputs to assess its iterative refinement behaviour and physical feasibility of the generated assemblies.

\begin{figure}[t] 
    \vspace{2mm}
    \centering
    \includegraphics[width=0.97\linewidth]{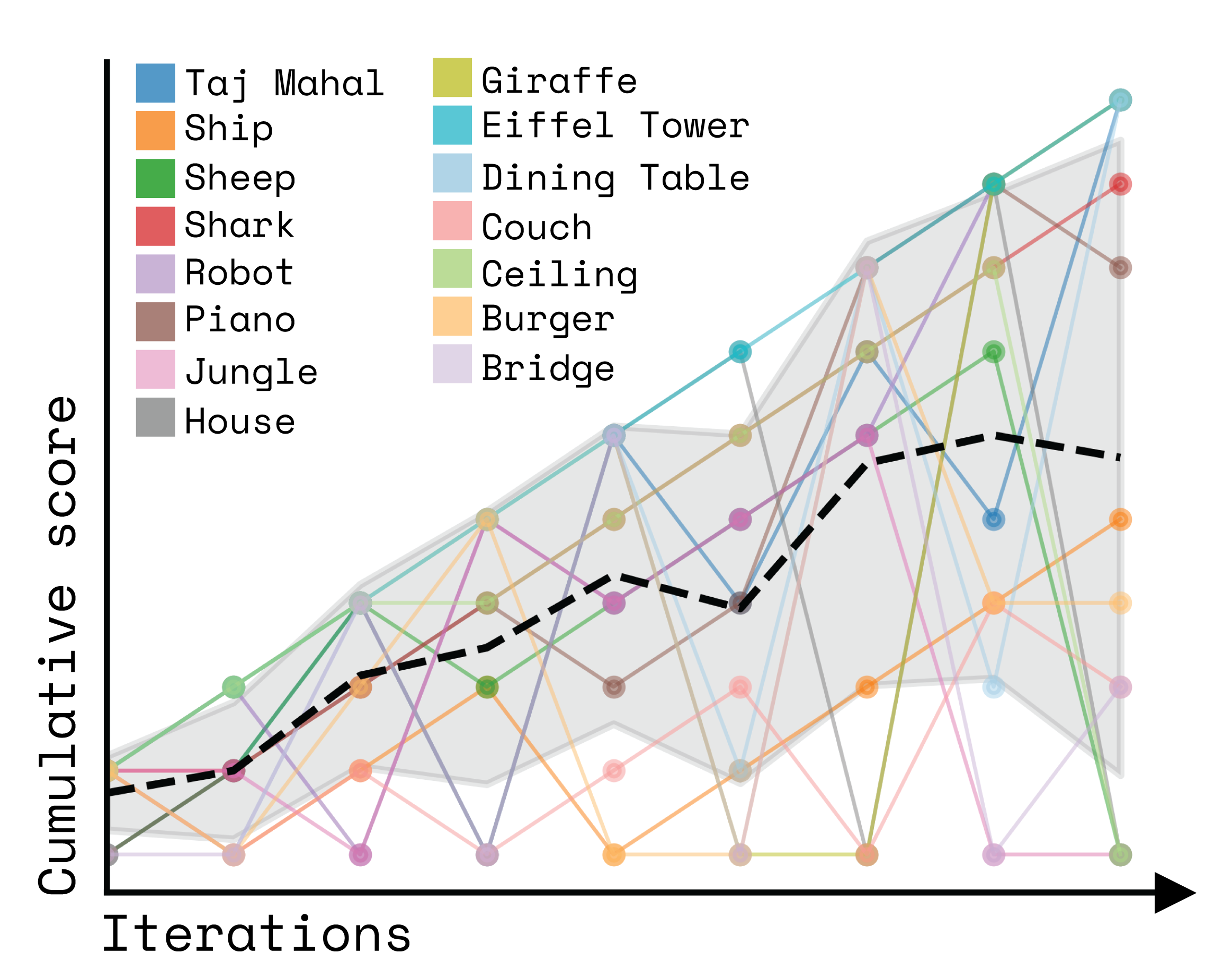}
    \caption{\textbf{Iterative design improvement}. For each structure, the Y-axis shows the number of times an iteration matched or outperformed previous iterations. The mean trend is indicated by a black dashed line with standard deviation shaded in grey.}
    \label{fig:iterative}
\end{figure}

\subsection{Human Evaluation Survey}
We conducted a human evaluation survey in which participants compared each pair of designs (IDfRA vs. BloxNet) for the 15 representative assemblies, and selected the one that more closely resembled the target structure. Order within each pair was randomised to mitigate bias. Participants were instructed to consider overall shape and key features, block usage, and stability. Recruitment combined convenience and snowball sampling.

Responses from 100 participants (Fig.~\ref{fig:human_eval}) show that IDfRA was preferred in 8 out of 15 designs, achieving a win rate of 54.2\% (813/1,500 pairwise comparisons). IDfRA outperformed BloxNet on more complex structures with multiple components such as ``robot" and ``Taj Mahal" (Fig~\ref{fig:qual}d), while BloxNet performed better on simpler targets such as ``ceiling fan" and ``bridge". In cases like ``Eiffel Tower" and ``burger", participants favoured BloxNet’s designs despite reduced stability and suboptimal block usage, suggesting that they prioritised overall shape over both of these. As in Fig~\ref{fig:qual}d, BloxNet’s "Eiffel Tower" was taller but less stable, and its "burger" achieved a flatter profile using a square block for the bun. These results reveal scope for adapting IDfRA’s design priorities in future work to reflect the dominant objectives of a given task, whether that be semantic resemblance, structural soundness, or accurate use of materials.

\subsection{VLM Test for Design Recognisability}
The second comparative assessment involved prompting a VLM with a rendered assembly image and an unordered list of objects, asking it to rank the objects based on semantic resemblance to the image. This approach, inspired by VLM-based answer scoring methods \cite{tvl} and previously employed by \cite{bloxnet}, was reproduced here to enable direct comparability with BloxNet. To avoid trivialising the task with unrelated distractors, we prompted the VLM to generate a list of 200 random objects including and semantically related to the 15 representative assemblies. For each of the 15 assemblies (IDfRA and BloxNet), we constructed ranking trials with list lengths $N = 5, 10, 15, 20$, where each unordered list contained the correct label and $N-1$ distractors. The VLM was then prompted to rank the objects by similarity to the rendered assembly image. Full object lists and code are included in the codebase.

As in \cite{bloxnet}, we report (1) Top-1 accuracy, the percentage of assemblies in which the correct label is ranked first, (2) Average Ranking, the mean ranking assigned to the correct label across all 15 assemblies, and (3) Relative Ranking, the mean rank of the correct label normalised by
the total number of choices $N$ in the list.

The results of this experiment are presented in Table~\ref{vlm_sem}. IDfRA consistently outperforms BloxNet in this evaluation, obtaining up to 73.33\% top-1 accuracy at $N = 10$. While IDfRA designs generally received higher rankings, both methods exhibited a common limitation: structures with similar high-level shapes were frequently confused -- “sheep” and “house” were often misclassified as ``table-like" structures such as “side table.” Additionally, certain structures disproportionately impacted performance. For example, ``shark” and ``piano” reduced accuracy for both methods, whereas “sheep” particularly reduced IDfRA’s scores, while ``Taj Mahal” degraded BloxNet's.

\begin{table}[t]
\vspace{2mm}
\centering
\renewcommand{\arraystretch}{1.2}
\resizebox{\columnwidth}{!}{%
\begin{tabular}{c l c c c}
\toprule
\textbf{N} & \textbf{Method} & \textbf{Top-1 Acc.} & \textbf{Average Ranking} & \textbf{Relative Ranking} \\
\midrule
\multirow{2}{*}{5} 
& Iterative DfRA & \textbf{64.44\%} & \textbf{1.60} & \textbf{32.00}\% \\
& BloxNet & 62.22\% & 1.62 & 32.44\% \\[2pt]
\hline
\multirow{2}{*}{10} 
& Iterative DfRA & \textbf{73.33\%} & \textbf{2.24} & \textbf{22.44\%} \\
& BloxNet & 57.78\% & 2.38 & 23.78\% \\[2pt]
\hline
\multirow{2}{*}{15} 
& Iterative DfRA & \textbf{68.89}\% & \textbf{2.71} & \textbf{18.07}\% \\
& BloxNet & 51.11\% & 3.36 & 22.37\% \\[2pt]
\hline
\multirow{2}{*}{20} 
& Iterative DfRA & \textbf{46.67\%} & \textbf{4.38} & \textbf{21.89}\% \\
& BloxNet & 42.22\% & 5.40 & 27.00\% \\
\bottomrule
\end{tabular}}
\caption{\textbf{VLM-based design recognisability}: Top-1 accuracy, average ranking, and relative ranking averaged across three runs. Computed using VLM rankings of N objects chosen from a pool of 200 objects based on semantic resemblance.}
\label{vlm_sem}
\end{table}

\begin{table}[t]
\centering
\renewcommand{\arraystretch}{1.2}
\resizebox{\columnwidth}{!}{%
\begin{tabular}{lcc}
\toprule
\textbf{Assembly Design} & \textbf{  \% Blocks Correctly Placed} & \textbf{  \% Assemblies Successful} \\
\midrule
Giraffe    & 97.78     & 80  \\
Taj Mahal  & 100    & 100 \\
House      & 100    & 100 \\
Shark       & 95    & 70 \\
Burger     & 100    & 100 \\
Jungle     & 95.71  & 70  \\
\bottomrule
\end{tabular}}
\caption{\textbf{Physical feasibility analysis}: Percentage of blocks correctly placed by the robot and percentage of successful assemblies across 10 trials for six assembly designs.}
\label{tab:exec_results}
\end{table}

\begin{table}[t]
\vspace{2mm}
\centering
\renewcommand{\arraystretch}{1.2}
\resizebox{\columnwidth}{!}{%
\begin{tabular}{p{2.3cm}ccccc}
\hline
 & \multicolumn{2}{c}{\textbf{VLM Recognisability (N = 10)}} & \multicolumn{2}{c}{\textbf{Containing Missing Blocks}} \\
 & \textbf{Top-1 Acc.} & \textbf{Average Rank} & \textbf{\# targets} & \textbf{\# iterations} \\
\midrule
Proposed IDfRA & \textbf{94.44}\% & \textbf{1.05} & \textbf{2} & 10 \\
PNG instead of GIF & 72.22\% & 1.61 & \textbf{2} & 9 \\
No missing blocks & 66.67\% & 2.22 & 3 & 11 \\
No history & 83.33\% & 2.05 & 4 & \textbf{5} \\
\bottomrule
\end{tabular}}
\caption{\textbf{Ablation studies}: VLM-based design recognisability and frequency of missing blocks across 10 iterations for each of the 6 representative targets (``bridge", ``burger", ``ceiling fan", ``couch", ``giraffe", ``robot"). }
\label{tab:ablations}
\end{table}

\subsection{Robot Assembly Test for Feasibility}
Next, we independently assessed the physical feasibility of IDfRA's generated designs within the simulation environment. For this analysis, we chose six assemblies, as in \cite{bloxnet}, to cover a range of block types, stacking heights, and difficulty levels. For each, the generated plans were executed in simulation over 10 trials, with the percentage of correctly placed blocks and fully successful build attempts reported in Table~\ref{tab:exec_results}. 

This evaluation aims to verify whether IDfRA’s designs are in principle constructible, despite using a minimal simulation environment; the setup prioritises design validation over precise robotic optimisation. The minimal simulation serves as a proxy for an industrial system, particularly in providing visual inputs to the \textit{Judge},  and offers a practical platform for proof of concept. 

As shown in Table~\ref{tab:exec_results}, IDfRA's designs are physically executable by the simulated robotic arm, with three out of six structures successfully assembled in all 10 trials; the remaining achieving at least 70\% success. Failures in ``giraffe” were attributed to the lack of collision checks, which caused the tail block to knock over the neck despite the design itself being valid (as evidenced by perfect execution in the other trials). ``Shark" and "jungle" exhibited unsuccessful trials due to isolated block placement errors, cuboids at the base edges requiring precise placement sometimes toppled during execution.

 Four generated designs spanning core manipulation challenges (e.g., stacking, edge placement), ``sheep”,
“jungle”, “burger”, and “Taj Mahal”, were also successfully assembled in a simple physical environment without plan modification, as shown in Fig~\ref{fig:physical}.

\subsection{Iterative Improvement Evaluation}
We examined how assembly designs evolved over ten iterations for each of the 15 target structures. For a given structure, each iteration was compared pairwise against all previous ones (e.g., iteration 3 vs. 2 and 1), yielding a binary outcome: 1 if the later design was better or equal, 0 otherwise. Each comparison was evaluated manually due to empirical limitations of VLMs in highly precise physical reasoning. If one design had  missing blocks, it automatically lost. If neither (or both) designs had missing blocks, the more stable design was preferred. For equally stable designs, the more semantically recognisable design won.

For each iteration $i$, we computed a cumulative score encoding the number of preceding iterations it matched or outperformed. These scores are plotted on the Y-axis of Fig.~\ref{fig:iterative} for iterations 1–9 (iteration 0 being the initial assembly). In this setup, a monotonically increasing line indicates a design that remains the same or improves over time.

Fig~\ref{fig:iterative} illustrates that designs remained the same or improved across iterations on average (black dashed line) with a greater standard deviation (shaded grey) in later iterations. Drops in the coloured lines correspond to unstable or incomplete assemblies, which received lower scores. Overall, the results confirm that designs remain the same or improve via iterative refinement, albeit not always monotonically.

\subsection{Ablation Studies}
We conducted three ablations: (1) using a PNG instead of GIF as visual input to the \textit{Judge}, (2) omitting missing block information from the \textit{Judge}, and (3) omitting plan history from the \textit{Replanner}. We analysed designs for 6 representative targets of varying complexity--``bridge", ``burger", ``ceiling fan", ``couch", ``giraffe", ``robot". We report VLM-based design recognisability (computed as in Table~\ref{tab:ablations}) and the number of targets (out of 6), as well as number of iterations (out of 6*10 = 60) containing missing blocks.

The proposed pipeline outperforms all ablations on semantic recognisability. While "No history" produces fewer total missing blocks (5 vs. 10), these failures occur across 4 targets compared to only 2 in the proposed method, demonstrating that the proposed approach concentrates failures in inherently challenging designs (``burger", ``robot") rather than causing sporadic failures across multiple targets (``bridge", ``burger", ``ceiling fan", ``giraffe").

\section{Conclusion}
\begin{figure}[t] 
\vspace{2mm}
    \centering
    \includegraphics[width=\linewidth]{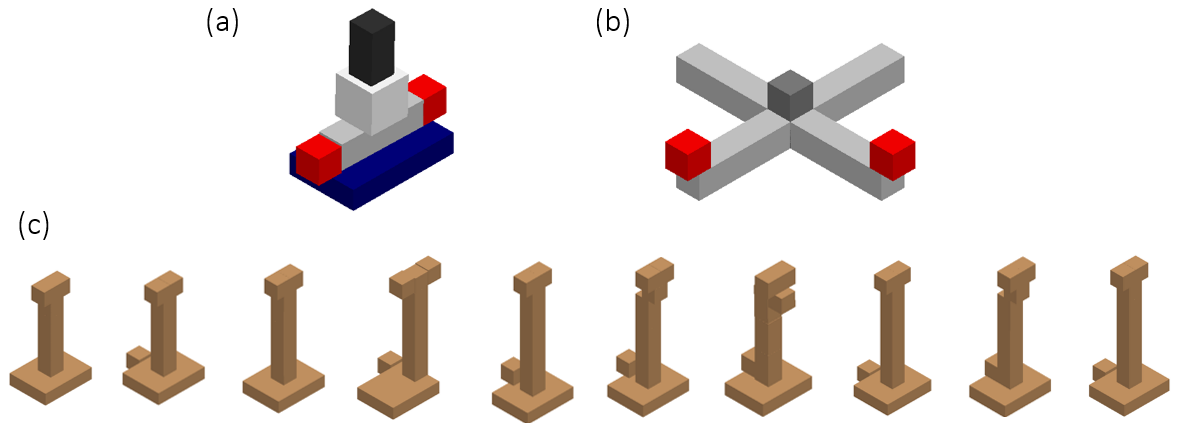}
    \caption{\textbf{Limitations}. IDfRA occasionally generates designs which contain overhanging blocks or include additional unnecessary details (highlighted in red) such as for ``ship" (a) and ``ceiling fan" (b). Though rare, it is also possible that no iteration produces a `good' design such as this run of ``giraffe" (c). Such failures largely stem from the VLM misidentifying unstable structures as stable.}
    \label{fig:limits}
\end{figure}

Experimental results demonstrate that IDfRA generates semantically recognisable and feasible assembly plans, and that iterative refinement can improve Design for Robotic Assembly (DfRA). The approach shows promise as a proof of concept, but several limitations leave scope for further exploration. The LLM-based \textit{Position} submodule remains prone to occasional inconsistencies despite prompt-engineering improvements (Fig~\ref{fig:limits}a, b), though overall performance is acceptable. Generated designs depend, to some extent, on the block set provided and are constrained by how well targets can be represented by a limited set of blocks. Future work could thus focus on scaling the system to more complex tasks and supporting larger, more diverse block sets in a real-world setting. The simulation environment was intentionally kept minimal to serve as a proxy for real robotic systems, with recorded GIFs mimicking real-world execution videos used in the planning pipeline. This design choice enables extension to physical robots and specialised manufacturing settings, where visual feedback plays a central role.

Finally, behaviour varies markedly across runs due to the inherent stochasticity of VLM generation. While this variability can produce diverse and creative “good” designs, it also introduces the rare risk that no satisfactory design emerges (Fig~\ref{fig:limits}c). Moreover, iterative refinement does not always yield monotonic improvement. Future work could address this by incorporating persistent memory, such as a “common mistakes to avoid” module or skill library \cite{voyager, srsa}, to encourage more consistent progress across iterations.

\begin{figure}[t] 
    \vspace{2mm}
    \centering
    \includegraphics[width=\linewidth]{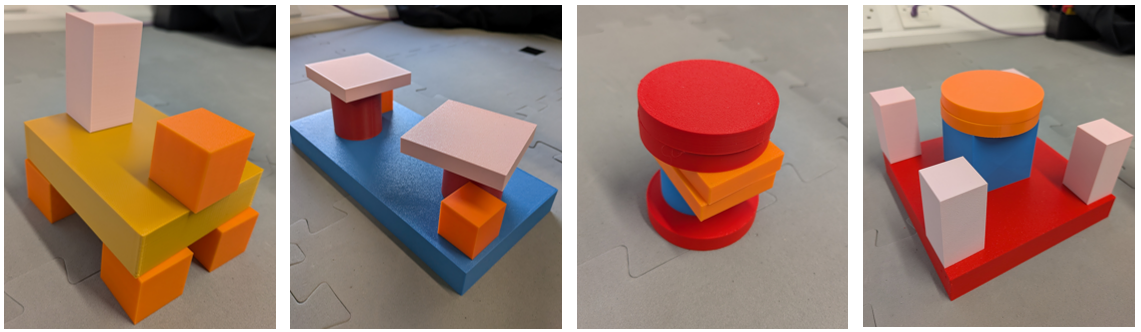}
    \caption{\textbf{Physical Experiments}. Four generated designs--``sheep", ``jungle", ``burger", and ``Taj Mahal"-- assembled using 3D printed blocks by an AGILEX PiPER arm and suction gripper for manipulation, and an Intel RealSense Depth Camera for localisation.}
    \label{fig:physical}
\end{figure}

Despite these limitations, IDfRA achieves Top-1 accuracy of up to 73.33\% in VLM-based object recognisability, and outperforms BloxNet in 8 out of 15 designs based on pairwise human evaluations--indicating enhanced semantic alignment for complex designs. The generated plans are also physically feasible, with an overall success rate of 86.67\% across six representative assemblies. As shown in Fig.~\ref{fig:physical}, the generated designs are directly executable by a real robotic system without plan modification, supporting transferability to real-world settings. Iterative evaluation further shows that designs often improve across successive refinements despite fluctuations. These findings highlight IDfRA’s potential as a self-refining, foundation-model-driven system capable of generating executable designs from natural language instructions. IDfRA holds particular promise for flexible robotic assembly in underspecified or evolving contexts, as it can dynamically adapt to feedback and environmental constraints through iterative self-verification and refinement.

\section*{ACKNOWLEDGMENT}
\noindent
SWP is a founder of Matta Labs, and CM is an employee; both hold equity in the company, which develops AI systems for manufacturing. The remaining author declares no competing interests.


\bibliographystyle{IEEEtran}
\bibliography{IEEEabrv,references}

\end{document}